\documentclass[10pt,twocolumn,letterpaper]{article}

\usepackage{cvpr}
\usepackage{times}
\usepackage{epsfig}
\usepackage{graphicx}
\usepackage{amsmath}
\usepackage{amssymb}
\usepackage{multirow}
\usepackage{subfigure}

\usepackage[breaklinks=true,bookmarks=false]{hyperref}

\cvprfinalcopy 

\begin{document}

\title{Semantically Consistent Image Completion with Fine-grained Details}

\author{Pengpeng Liu, Xiaojuan Qi, Pinjia He, Yikang Li, Michael R. Lyu, Irwin King\\
The Chinese University of Hong Kong, Hong Kong SAR, China\\
{\tt\small \{ppliu, xjqi, pjhe, ykli, lyu, king\}@cse.cuhk.edu.hk}
}

\maketitle

\begin{abstract}
Image completion has achieved significant progress due to advances in generative adversarial networks (GANs). Albeit natural-looking, the synthesized contents still lack details, especially for scenes with complex structures or images with large holes. This is because there exists a gap between low-level reconstruction loss and high-level adversarial loss. To address this issue, we introduce a perceptual network to provide mid-level guidance, which measures the semantical similarity between the synthesized and original contents in a similarity-enhanced space. We conduct a detailed analysis on the effects of different losses and different levels of perceptual features in image completion, showing that there exist complementarity between adversarial training and perceptual features. By combining them together, our model can achieve nearly seamless fusion results in an end-to-end manner. Moreover, we design an effective lightweight generator architecture, which can achieve effective image inpainting with far less parameters. Evaluated on CelebA Face and Paris StreetView dataset, our proposed method significantly outperforms existing methods.
\end{abstract}

\section{Introduction}
\begin{figure}
\centering
\subfigure[Original]{
\begin{minipage}[b]{0.15\textwidth}
\includegraphics[width=1\textwidth]{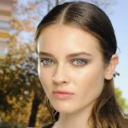}  \\ 
\includegraphics[width=1\textwidth]{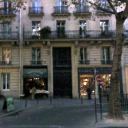} 
\end{minipage}
}
\hspace{-0.015\textwidth}
\subfigure[Masked]{
\begin{minipage}[b]{0.15\textwidth}
\includegraphics[width=1\textwidth]{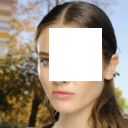} \\
\includegraphics[width=1\textwidth]{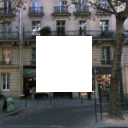} 
\end{minipage}
}
\hspace{-0.015\textwidth}
\subfigure[Results]{
\begin{minipage}[b]{0.15\textwidth}
\includegraphics[width=1\textwidth]{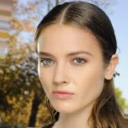} \\
\includegraphics[width=1\textwidth]{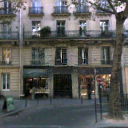} 
\end{minipage}
}
\caption{Image completion results by our method. Note that the synthesized contents are different from the original images, but the completion results still look natural and semantically consistent.}
\label{Samples}
\end{figure}

Image completion, also known as image inpainting, aims to restore the damaged images or fill in the missing parts of images with visually plausible contents (Figure~\ref{Samples}). As a common image editing technique, it can also be used to remove unwanted objects. Traditional image completion methods \cite{bertalmio2000image,elad2005simultaneous,bertalmio2003simultaneous,barnes2009patchmatch} utilize low-level information to synthesize missing regions, which fails to produce high-level image semantics. By stacking multiple layers, CNNs are able to capture some intrinsic hierarchical representations. They have been employed in this field and achieve great success \cite{xie2012image,kohler2014mask}. However, most of them are designed for completing small and narrow holes, such as removing text in images. 

\begin{figure*}
\centering
\subfigure[Original]{
\begin{minipage}[t]{0.15\textwidth}
\includegraphics[width=1\textwidth]{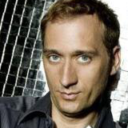} \\
\includegraphics[width=1\textwidth]{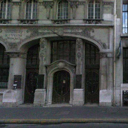} 
\end{minipage}
\label{comparision: a}
}
\hspace{-0.015\textwidth}
\subfigure[Masked]{
\begin{minipage}[t]{0.15\textwidth}
\includegraphics[width=1\textwidth]{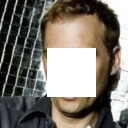} \\
\includegraphics[width=1\textwidth]{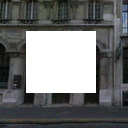} 
\end{minipage}
\label{comparision: b}
}
\hspace{-0.015\textwidth}
\subfigure[$L_r$]{
\begin{minipage}[t]{0.15\textwidth}
\includegraphics[width=1\textwidth]{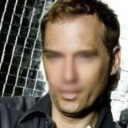} \\
\includegraphics[width=1\textwidth]{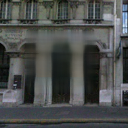} 
\end{minipage}
\label{comparision: c}
}
\hspace{-0.015\textwidth}
\subfigure[$L_r+L_a$]{
\begin{minipage}[t]{0.15\textwidth}
\includegraphics[width=1\textwidth]{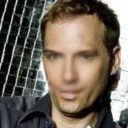} \\
\includegraphics[width=1\textwidth]{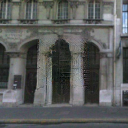} 
\end{minipage}
\label{comparision: d}
}
\hspace{-0.015\textwidth}
\subfigure[$L_r+L_p$]{
\begin{minipage}[t]{0.15\textwidth}
\includegraphics[width=1\textwidth]{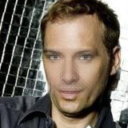} \\
\includegraphics[width=1\textwidth]{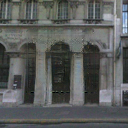} 
\end{minipage}
\label{comparision: e}
}
\hspace{-0.015\textwidth}
\subfigure[$L_r+L_a+L_p$]{
\begin{minipage}[t]{0.15\textwidth}
\includegraphics[width=1\textwidth]{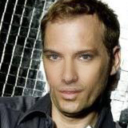} \\
\includegraphics[width=1\textwidth]{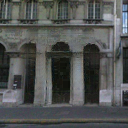} 
\end{minipage}
\label{comparision: f}
}
\caption{Our image completion results under different settings. (c) Using reconstruction loss $L_r$ only, completion results are very blurry; (d) Only adding adversarial loss $L_a$, completion results are a litter sharper and natural, but the synthesized contents still lack details; (e) Only adding perceptual loss $L_p$, results contain more details, but have clear artifacts and are discontinuous on the boundary (such as color difference); (f) The combination of them produce semantically consistent results with details. (f) shows our final results without any post-precessing procedures.}
\label{comparison}
\end{figure*}

Recently proposed adversarial training is a powerful strategy to train an image inpainting model~\cite{pathak2016context,yeh2016semantic,li2017generative}. During training, two loss functions are widely used, the pixel-wise reconstruction loss and image-level adversarial loss. The former one helps the model reconstruct the original missing part by ensuring the pixel-wise identity~(too low-level), while the latter aims at making the completed image natural-looking enough to fool the discriminator (too high-level). Intuitively, a mid-level guidance, which can encourage the restoration of differently-looking but semantically consistent contents, is lacking.

To close the gap, we propose to utilize a pre-trained CNN~(termed as \emph{perceptual network}) in our framework to transform the original image into a suitable feature space, where semantical similarity is enhanced instead of pixel-wise appearance. 
In this space, differently-looking but semantically consistent contents will get closer. 
By emphasizing the similarity between the contents and the original contents on the feature level, an additional loss is proposed, termed as \emph{Perceptual Loss}. By minimizing perceptual loss, the model gets more encouragement on generating differently-looking but semantically consistent contents.

In practice, we find that adversarial training and perceptual features are complementary to each other, which is consistent with our analysis that they provide different-level semantic supervision. Adversarial training helps to make the completed image natural and enables the synthesized contents continuous on the boundary, but the synthesized contents usually lack details (Figure~\ref{comparision: d}). Perceptual features can guide the generator to synthesize fine-grained details. However, they bring artifacts and the synthesized contents are discontinuous with the surrounding regions on the boundary of missing regions (Figure~\ref{comparision: e}). Thus, by combining them together, our proposed method can produce consistent contents with fine-grained details (Figure~\ref{comparision: f}). Since different perceptual features have different effects on contents synthesis, we further make a detailed analysis and comparison of different perceptual features (Figure~\ref{Perceptual comparison}). 

Besides, most existing learning-based image completion approaches leverage extremely large models (e.g., number of parameters \textgreater 100M) to produce reasonable results, which does not fit for real application. Thus, we design a novel generator structure based on fully convolutional network (FCN), which can achieve better performance with far less parameters (number of parameters \textless 10M).  

Our contributions can be summarized as follows. We propose to employ supervision of multiple levels for image completion: reconstruction loss (low-level), perceptual loss (mid-level) and adversarial loss (high-level). Based on the detailed analysis and comparison on the effects of different loss functions and different levels of perceptual features, we show that there exists complementarity between adversarial training and perceptual features. Besides, a novel lightweight FCN-based generator structure is designed to produce better inpainting results with far less parameters. Experiments on CelebA Face and Paris StreetView demonstrates the superiority of out proposed method over the existing methods.

\begin{figure*}[htbp]
\centering
\includegraphics[width=16cm]{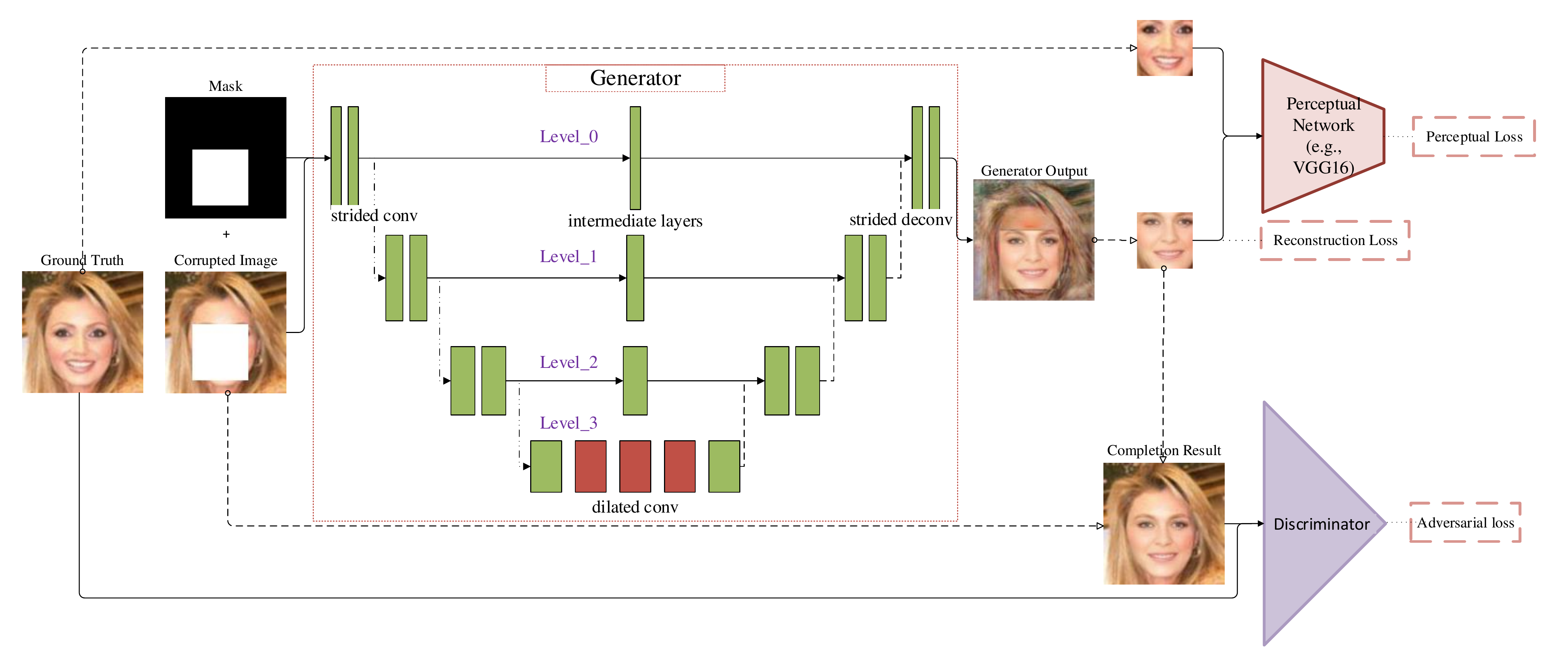}
\caption{Framework Overview. The generator in our framework can have different number of levels, and it contains three levels in this example. Higher resolution images need more levels to capture larger receptive fields. The discriminator helps make the completion results natural as a whole and enables the boundary continuity. The perceptual network can capture different semantic-level features, which guides the generator to synthesize detailed contents. Only the generator is needed during testing stage.}
\label{ModelStrcuture}
\end{figure*}

\section{Related Work}
\subsection{Adversarial Training} 

Generative Adversarial Network (GAN) \cite{goodfellow2014generative} is a minimax two-player game, which utilizes adversarial training to train the generator and discriminator alternatively and has shown powerful ability to generate natural and high-quality images. Denton~\etal~\cite{denton2015deep} applies a Laplacian pyramid GAN framework to generate high-resolution images in a coarse-to-fine manner. Radford~\etal~\cite{radford2015unsupervised} proposes deep convolutional GAN (DCGAN), which generates high-quality images in many datasets. Arjovsky~\etal~\cite{arjovsky2017wasserstein} analyzes the causes of instability theoretically and puts forward Wasserstein GAN.

Although GAN is firstly proposed for image generation, the idea of adversarial training is actually general, which has been widely applied to various research fields, such as image-to-image translation \cite{isola2016image} and image suer-resolution \cite{ledig2016photo,sonderby2016amortised}. Different from these existing works, we apply adversarial training to generate natural image completion results and ensure the continuity between the synthesized contents and the surrounding regions.
\subsection{Perceptual Features} 
Perceptual features, which are extracted from pre-trained networks (e.g., VGG16 \cite{simonyan2014very}), have been employed in many computer vision tasks. Gatys~\etal~\cite{gatys2016image} applies perceptual features to neural style transfer, which recombines one image's content and another image's style for high-level image synthesis. Johnson~\etal~\cite{johnson2016perceptual} trains a feed-forward neural network to solve the optimization problem and improves its speed to achieve real-time style transfer. Ledig~\etal~\cite{ledig2016photo} make use of perceptual features for realistic image super-resolution. Since perceptual features can capture both low-level and high-level semantic information, Dosovitskiy~\etal~\cite{dosovitskiy2016generating} utilizes it to generate natural images. We show that perceptual features can also be used for generating details in image completion.

\subsection{Image Completion}
Traditional image completion algorithms can be broadly divided into two categories. The first category is diffusion-based \cite{bertalmio2000image,elad2005simultaneous}, which propagates low-level features from surrounding regions to the missing holes. However, this category cannot synthesize pleasing contents for textured regions or large holes. The second category is patch-based, which searches similar patches from the same image or database \cite{bertalmio2003simultaneous,barnes2009patchmatch}. Methods in this category only make use of low-level features to ensure local similarity and will fail if similar patches do not exist in the database.

Recently, Context Encoder \cite{pathak2016context} produces reasonable semantic image inpainting results for the first time. It trains an auto-encoder neural network to predict the missing regions with the combination of $L_2$ loss and adversarial loss. Yeh~\etal~\cite{yeh2016semantic} considers image completion as an image generation problem, which leverages a pre-trained DCGAN \cite{radford2015unsupervised} model and tries to find the closest vector in the latent space. However, it needs many iterations before finding a proper latent vector during testing stage. Moreover, pre-trained GAN models do not always exist. Li~\etal~\cite{li2017generative} proposes a generative approach for face completion, which utilizes two adversarial losses and can generate realistic face completion results. However, it cannot work well on misaligned face images and the need for a face scene parsing network  restricts its application in other scenes. These large hole-filling approaches only take advantage of adversarial training, thus the inpainting results usually lack high-frequency details and have obvious color difference along mask boundaries, needing some post-processing methods such as Poisson blending \cite{perez2003poisson} to eliminate. Our algorithm can synthesize semantically consistent contents with details and requires no post-processing procedures.

\begin{figure*}
\centering
\subfigure[Masked]{
\begin{minipage}[t]{0.13\textwidth}
\includegraphics[width=1\textwidth]{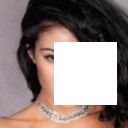} \\
\includegraphics[width=1\textwidth]{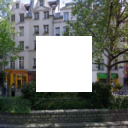} 
\end{minipage}
\label{Perceptual comparison: a}
}
\hspace{-0.015\textwidth}
\subfigure[$L_r+L_{p_1}$]{
\begin{minipage}[t]{0.13\textwidth}
\includegraphics[width=1\textwidth]{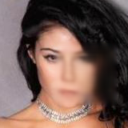} \\
\includegraphics[width=1\textwidth]{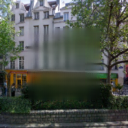} 
\end{minipage}
\label{Perceptual comparison: b}
}
\hspace{-0.015\textwidth}
\subfigure[$L_r+L_{p_2}$]{
\begin{minipage}[t]{0.13\textwidth}
\includegraphics[width=1\textwidth]{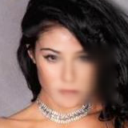} \\
\includegraphics[width=1\textwidth]{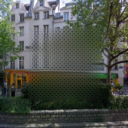} 
\end{minipage}
\label{Perceptual comparison: c}
}
\hspace{-0.015\textwidth}
\subfigure[$L_r+L_{p_3}$]{
\begin{minipage}[t]{0.13\textwidth}
\includegraphics[width=1\textwidth]{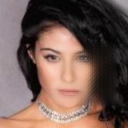} \\
\includegraphics[width=1\textwidth]{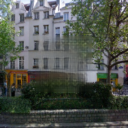} 
\end{minipage}
\label{Perceptual comparison: d}
}
\hspace{-0.015\textwidth}
\subfigure[$L_r+L_{p_4}$]{
\begin{minipage}[t]{0.13\textwidth}
\includegraphics[width=1\textwidth]{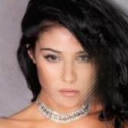} \\
\includegraphics[width=1\textwidth]{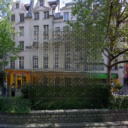} 
\end{minipage}
\label{Perceptual comparison: e}
}
\hspace{-0.015\textwidth}
\subfigure[$L_r+L_{p_5}$]{
\begin{minipage}[t]{0.13\textwidth}
\includegraphics[width=1\textwidth]{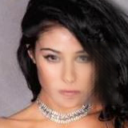} \\
\includegraphics[width=1\textwidth]{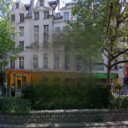} 
\end{minipage}
\label{Perceptual comparison: f}
}
\hspace{-0.015\textwidth}
\subfigure[$L_r+\{L_{p_i}\}$]{
\begin{minipage}[t]{0.13\textwidth}
\includegraphics[width=1\textwidth]{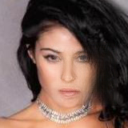} \\
\includegraphics[width=1\textwidth]{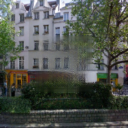} 
\end{minipage}
\label{Perceptual comparison: g}
}
\caption{The effect of different perceptual features. $L_{p_1}$ is computed based on the output of $conv1\_2$ in VGG16. Similarily, from (c) to (f), we use perceptual features of $conv2\_2$, $conv3\_2$, $conv4\_2$, $conv5\_2$. $\{L_{p_i}\}$ means the combination of several perceptual losses. (b)(c) produce blurry results, meaning that $conv1\_2$ and $conv2\_2$ only capture low-level features. (d)(e)(f) contain more details because they capture some semantic-level representations. (g) is the combination of several semantic-level features and produces the best results. }
\label{Perceptual comparison}
\end{figure*}
\section{Proposed Method}
Figure~\ref{ModelStrcuture} shows the overall architecture of our framework. The generator is used to synthesize the missing contents. The discriminator makes the completed image natural as a whole and enables the boundary continuity between the synthesized contents and the surrounding regions. The perceptual network can capture mid-level semantic features, which guide the generator to synthesize detailed contents.

\subsection{Generator}
The input of our generator is the corrupted image plus a mask indicating which pixels are missing. Reconstruction loss and perceptual loss are only applied to the masked regions, because we do not care about the unmasked regions of generator output (`Generator Output' in Figure~\ref{ModelStrcuture}). Even so, the generator output is an image of the same resolution as original images, which enables our algorithm to handle the cases of random mask shape and location. 

Our generator structure is based on an FCN, which makes use of two good properties of convolution neural networks: translation invariance and parameter sharing. The former is essentially helpful for image completion of arbitrary mask location, and the latter can reduce the number of parameters tremendously. However, since convolution layers are locally connected, some input pixels cannot influence output pixels if the receptive field is small (e.g., receptive field size $<$ hole size).  

Stacking many layers or downsampling can enlarge receptive field. However, the former increases the number of parameters and computation cost, and the latter decreases image restoration performance \cite{mao2016image}. To address the above issues, we employ dilated convolution layers \cite{yu2015multi} to enlarge receptive field without increasing the number of parameters. As shown in Figure~\ref{ModelStrcuture}, we call each downsampling as a $level$, and stack several dilated convolution layers in the last level. Besides, we propose to add intermediate layers at different levels to combine multi-scale information. With the large receptive field and multi-scale information fusion, our generator structure can achieve comparable performance with far less parameters (e.g., set max channel number as 128).

In our implementation, we use $3 \times 3$ kernel size in all convolution layers, $4 \times 4$ kernel size with stride 2 in deconvolution layers \cite{long2015fully}. Apart from the last layer, we add batch normalization (BN) \cite{ioffe2015batch} and ReLU activation function after convolution. Different generator structures are compared in the experiment part (Figure \ref{Generator Compare}), which demonstrates the superiority of our generator. 

\subsection{Discriminator}
Reconstruction loss tends to average all of the details, thus the generator can only capture coarse information about the missing regions. Therefore the synthesized contents look blurry (Figure~\ref{comparision: c}). Besides, the generator only optimize the masked regions, thus the synthesized content can be discontinuous with surroundings. 

\begin{figure}
\centering
\subfigure[original]{
\begin{minipage}[b]{0.11\textwidth}
\includegraphics[width=1\textwidth]{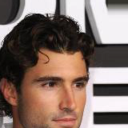}  \\ 
\includegraphics[width=1\textwidth]{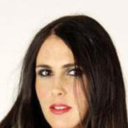}  \\
\includegraphics[width=1\textwidth]{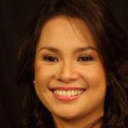}
\end{minipage}
\label{Replace Compare: a}
}
\hspace{-0.015\textwidth}
\subfigure[masked]{
\begin{minipage}[b]{0.11\textwidth}
\includegraphics[width=1\textwidth]{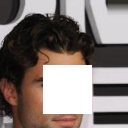}  \\ 
\includegraphics[width=1\textwidth]{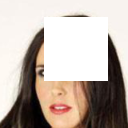}  \\
\includegraphics[width=1\textwidth]{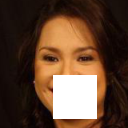}
\end{minipage}
\label{Replace Compare: b}
}
\hspace{-0.015\textwidth}
\subfigure[no replace]{
\begin{minipage}[b]{0.11\textwidth}
\includegraphics[width=1\textwidth]{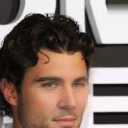}  \\ 
\includegraphics[width=1\textwidth]{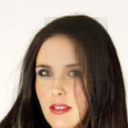}  \\
\includegraphics[width=1\textwidth]{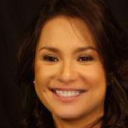}
\end{minipage}
\label{Replace Compare: c}
}
\hspace{-0.015\textwidth}
\subfigure[replace]{
\begin{minipage}[b]{0.11\textwidth}
\includegraphics[width=1\textwidth]{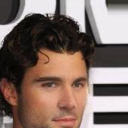}  \\ 
\includegraphics[width=1\textwidth]{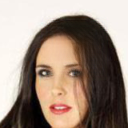}  \\
\includegraphics[width=1\textwidth]{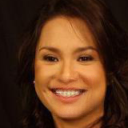}
\end{minipage}
\label{Replace Compare: d}
}
\caption{The effect of replacing unmasked regions with original pixels.}
\label{Replace Compare}
\end{figure}

Adversarial training trains discriminator and generator alternatively, which is known to be able to generate natural and realistic images. Therefore we employ a discriminator to distinguish between the original image and the completed image. The unmasked regions are replaced by the original image (`Completion Result' in Figure~\ref{ModelStrcuture}). This is essential and intuitive because the raw generator output (`Generator Output' in Figure~\ref{ModelStrcuture}) is not natural at all and discontinuous on the boundary. The effect of original image context can be seen in Figure~\ref{Replace Compare: c} and \ref{Replace Compare: d}.  

As shown in Figure~\ref{comparision: c}, \ref{comparision: d} and Figure~\ref{comparision: e}, \ref{comparision: f}, adding adversarial training can indeed make the result more natural and continuous on the boundary between the synthesized contents and surrounding regions.
 
Our discriminator structure is similar to DCGAN, using BN and Leakly ReLU activation function after each convolution layer. 
\begin{figure*}[ht]
\centering
\subfigure[original]{
\begin{minipage}[t]{0.13\textwidth}
\includegraphics[width=1\textwidth]{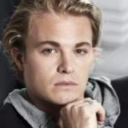} \\
\includegraphics[width=1\textwidth]{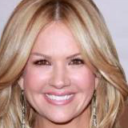} \\
\includegraphics[width=1\textwidth]{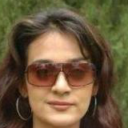} \\
\includegraphics[width=1\textwidth]{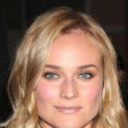} 
\end{minipage}
\label{CelebA Compare: a}
}
\hspace{-0.015\textwidth}
\subfigure[masked]{
\begin{minipage}[t]{0.13\textwidth}
\includegraphics[width=1\textwidth]{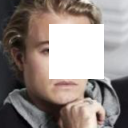} \\
\includegraphics[width=1\textwidth]{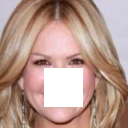} \\
\includegraphics[width=1\textwidth]{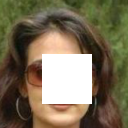} \\
\includegraphics[width=1\textwidth]{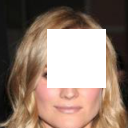} 
\end{minipage}
\label{CelebA Compare: b}
}
\hspace{-0.015\textwidth}
\subfigure[CAF]{
\begin{minipage}[t]{0.13\textwidth}
\includegraphics[width=1\textwidth]{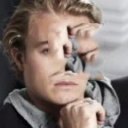} \\
\includegraphics[width=1\textwidth]{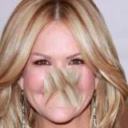} \\
\includegraphics[width=1\textwidth]{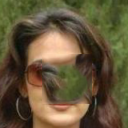} \\
\includegraphics[width=1\textwidth]{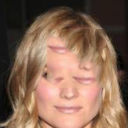} 
\end{minipage}
\label{CelebA Compare: c}
}
\hspace{-0.015\textwidth}
\subfigure[GFC]{
\begin{minipage}[t]{0.13\textwidth}
\includegraphics[width=1\textwidth]{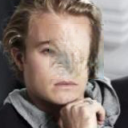} \\
\includegraphics[width=1\textwidth]{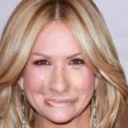} \\
\includegraphics[width=1\textwidth]{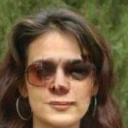} \\
\includegraphics[width=1\textwidth]{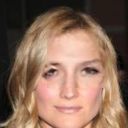} 
\end{minipage}
\label{CelebA Compare: d}
}
\hspace{-0.015\textwidth}
\subfigure[GFC\_post]{
\begin{minipage}[t]{0.13\textwidth}
\includegraphics[width=1\textwidth]{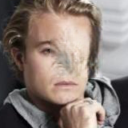} \\
\includegraphics[width=1\textwidth]{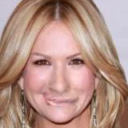} \\
\includegraphics[width=1\textwidth]{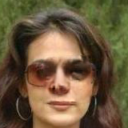} \\
\includegraphics[width=1\textwidth]{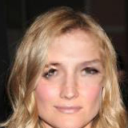}
\end{minipage}
\label{CelebA Compare: e}
}
\hspace{-0.015\textwidth}
\subfigure[Ours($L_r+L_a$)]{
\begin{minipage}[t]{0.13\textwidth}
\includegraphics[width=1\textwidth]{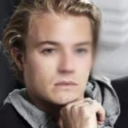} \\
\includegraphics[width=1\textwidth]{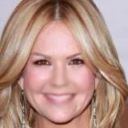} \\
\includegraphics[width=1\textwidth]{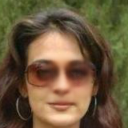} \\
\includegraphics[width=1\textwidth]{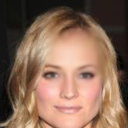} 
\end{minipage}
\label{CelebA Compare: f}
}
\hspace{-0.015\textwidth}
\subfigure[Ours(final)]{
\begin{minipage}[t]{0.13\textwidth}
\includegraphics[width=1\textwidth]{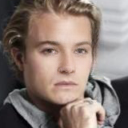} \\
\includegraphics[width=1\textwidth]{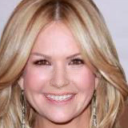} \\
\includegraphics[width=1\textwidth]{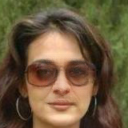} \\
\includegraphics[width=1\textwidth]{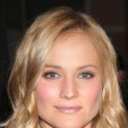} 
\end{minipage}
\label{CelebA Compare: g}
}
\caption{Comparison with Content Aware Fill (CAF), Generative Face Completion (GFC) and GFC with post-processing (GFC\_post) on CelebA. Our approach can produce more natural results, such as smile in the second row, and can synthesize more details, such as eyes in first row, teeth in second row, glasses in third row and hair in last row.} 
\label{CelebA Compare}
\end{figure*}
\subsection{Perceptual Network}
Discriminator can help produce shaper results, but as shown in Figure~\ref{comparision: d}, the synthesized contents still lack details. This is because discriminator only enforces the completion results to look natural as a whole and ignores some essential information for detailed contents synthesis. Perceptual features can capture different semantic-levels features, and by employing semantical similarity between the contents and the original contents on the feature level, the model gets more encouragement on generating semantically consistent contents with fine-grained details. 

However, only adding perceptual loss is not a good choice as well, which will lead to artifacts (Figure~\ref{comparision: e}). The reason is that there may exist many unnatural contents sharing the same representation in feature space. That is, similar in feature space does not ensure natural in image space. Therefore, we use both discriminator and perceptual network in our proposed framework.
\begin{figure*}[!htbp]
\centering
\subfigure[original]{
\begin{minipage}[b]{0.13\textwidth}
\includegraphics[width=1\textwidth]{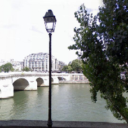} \\
\includegraphics[width=1\textwidth]{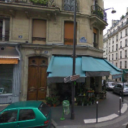} \\
\includegraphics[width=1\textwidth]{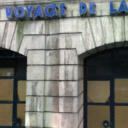} \\
\includegraphics[width=1\textwidth]{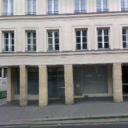}
\end{minipage}
\label{Paris Compare: a}
}
\hspace{-0.015\textwidth}
\subfigure[masked]{
\begin{minipage}[b]{0.13\textwidth}
\includegraphics[width=1\textwidth]{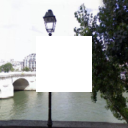} \\ 
\includegraphics[width=1\textwidth]{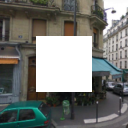} \\
\includegraphics[width=1\textwidth]{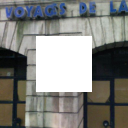} \\
\includegraphics[width=1\textwidth]{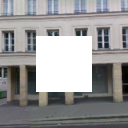} 
\end{minipage}
\label{Paris Compare: b}
}
\hspace{-0.015\textwidth}
\subfigure[CAF]{
\begin{minipage}[b]{0.13\textwidth}
\includegraphics[width=1\textwidth]{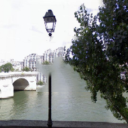} \\ 
\includegraphics[width=1\textwidth]{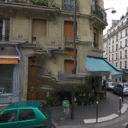} \\ 
\includegraphics[width=1\textwidth]{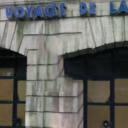} \\
\includegraphics[width=1\textwidth]{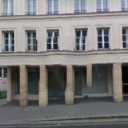}
\end{minipage}
\label{Paris Compare: c}
}
\hspace{-0.015\textwidth}
\subfigure[CE]{
\begin{minipage}[b]{0.13\textwidth}
\includegraphics[width=1\textwidth]{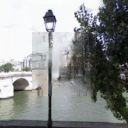} \\
\includegraphics[width=1\textwidth]{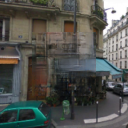} \\
\includegraphics[width=1\textwidth]{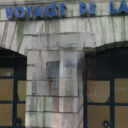} \\
\includegraphics[width=1\textwidth]{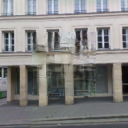} 
\end{minipage}
\label{Paris Compare: d}
}
\hspace{-0.015\textwidth}
\subfigure[Ours($L_r+L_a$)]{
\begin{minipage}[b]{0.13\textwidth}
\includegraphics[width=1\textwidth]{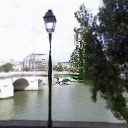} \\
\includegraphics[width=1\textwidth]{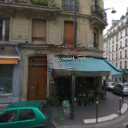} \\
\includegraphics[width=1\textwidth]{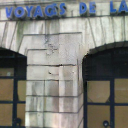} \\
\includegraphics[width=1\textwidth]{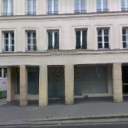}
\end{minipage}
\label{Paris Compare: e}
}
\hspace{-0.015\textwidth}
\subfigure[Ours(final)]{
\begin{minipage}[b]{0.13\textwidth}
\includegraphics[width=1\textwidth]{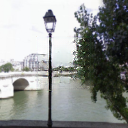} \\
\includegraphics[width=1\textwidth]{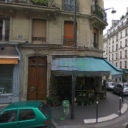} \\
\includegraphics[width=1\textwidth]{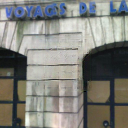} \\
\includegraphics[width=1\textwidth]{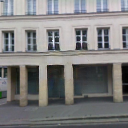}
\end{minipage}
\label{Paris Compare: f}
}
\caption{Comparison with Content Aware Fill (CAF), Context Encoder (CE) on Paris StreetView. Our method can handle different kinds of masks, but CE only works for specific masks used during training stage. Thus here we follow the original paper of CE and compare inpainting results on the central holes. }
\label{Paris Compare}
\end{figure*}

\subsection{Loss Functions}
We train the model with a special designed hybrid loss $L$, which is the combination of reconstruction loss $L_r$, adversarial loss $L_{g}$ and perceptual loss $L_{p}$. As shown in Figure~\ref{ModelStrcuture}, they correspond to generator, discriminator and perceptual network respectively. The overall loss function is defined as follow:
\begin{equation}
L=  L_{r} + \lambda_{1}L_{a} + \lambda_{2}L_{p},
\end{equation}
where $\lambda_{1}$ and $\lambda_{2}$ are hyperparameters that balance the contribution of different losses. 

We denote $I_{gt}$, $I_{c}$, $I_{g}$ as the original, corrupted and generated image respectively. Let $M$ be the binary mask, which has value 1 if the corresponding pixel needs to be completed and 0 otherwise.  

\textbf{Reconstruction Loss.} We utilize a smooth loss function $L_{smooth}$ to make an elegant compromise between $L_{1}$ and $L_{2}$, which is defined as:

\begin{equation} \begin{aligned}
     L_{smooth}(x) &= \left\{ \begin{array}{ll}
            0.5 x^2, & \textrm{if $|x| < 1$}\\
             |x|-0.5, & \textrm{otherwise}
                    \end{array} \right.
    \end{aligned}
\end{equation}

Our reconstruction loss is defined on the masked regions:
\begin{equation}
L_{r}= L_{smooth}(I_g \odot {M} - I_{gt} \odot {M}),
\end{equation}
where $\odot$ denotes the element-wise multiplication.

\textbf{Adversarial Loss.} Adversarial training trains generator and discriminator alternatively. The discriminator is trained to distinguish whether the corrupted image is completed naturally or not, while the generator tries to fool the discriminator. The image completion result is the combination of the masked regions of $I_g$ and unmasked regions of $I_{gt}$:
\begin{equation}
I_{completion}= M \odot I_g + (1-M) \odot I_{gt},
\end{equation}
therefore we can obtain the adversarial loss as follows:
\begin{equation}
L_{a}^D = -(\mathbb{E}{ [\log D(I_{gt})]} + \mathbb{E} [\log (1 - D(I_{completion})]),
\end{equation}
\begin{equation}
L_{a}^G = -\mathbb{E}[\log  D(I_{completion})].
\end{equation}
To stabilize the adversarial training, we adopt one-sided label smoothing strategy \cite{salimans2016improved}, which replaces the label 0 and 1 for a classifier with smoothed values. 
We add smoothed values on $L^D {adv}$, using a random values between 0 and 0.2 to replace label 0, another random values between 0.8 and 1 to replace label 1.

\textbf{Perceptual Loss.} Perceptual loss is based on perceptual network and different layers can capture different levels of features. Let $\{\phi_{l}\}$ be a collection of layers in the pre-trained perceptual network. The Perceptual loss function is defined as follows:
\begin{equation}
L_{p} = \sum_{l}\alpha_{l}L_{smooth}(\phi_{l}(I_{gt} \odot {M}) - \phi_{l}(G \odot {M})),
\end{equation}
where $\alpha_{l}$ is the weighted hyperparameter of layer $l$.

\section{Experiments}
In this section, we first introduce datasets and our experimental settings. Then we conduct four main experiments, which demonstrate the effectiveness of our image completion framework: comparison of different generator structures, effect of original image context, analysis and comparison of different levels of perceptual features, qualitatively and quantitatively evaluation of our image completion results. Finally, we show some real-word applications, such as object removal.

\subsection{Datasets}
We evaluate our method on CelebA Face dataset \cite{liu2015deep} and Paris StreetView dataset \cite{doersch2012makes}. CelebA contains 202,599 face images. We randomly choose 200,000 for training, the other 2,599 for testing. Paris StreetView contains 15,000 street scenes in Paris. As in Context Encoder \cite{pathak2016context}, 14,900 images are used for training and 100 for testing.

\subsection{Experimental Settings}
We implement our model on TensorFlow \cite{abadi2016tensorflow} and use Adam \cite{kingma2014adam} for optimization.

\textbf{Data Preprocessing.}
For CelebA, we randomly crop a 160 $\times$ 160 region and resize it to 128 $\times$ 128. For Paris StreetView, we first resize the image so that the smaller dimension is with size 128 and then 128 $\times$ 128 is cropped. We also conduct data augmentation such as random flipping, shift. We randomly choose mask size ranging from 48 to 80 and the mask location is totally random. Such random setting enables our model to complete image with arbitrary mask size and location. After choosing the mask size and location, we shift and rescale the pixel value from [0, 255] to [-1, 1] and fill in the masked regions with 0. Finally, the corrupted image and the mask are concatenated as input. 
\begin{figure}[t]
\centering
\includegraphics[width=8cm]{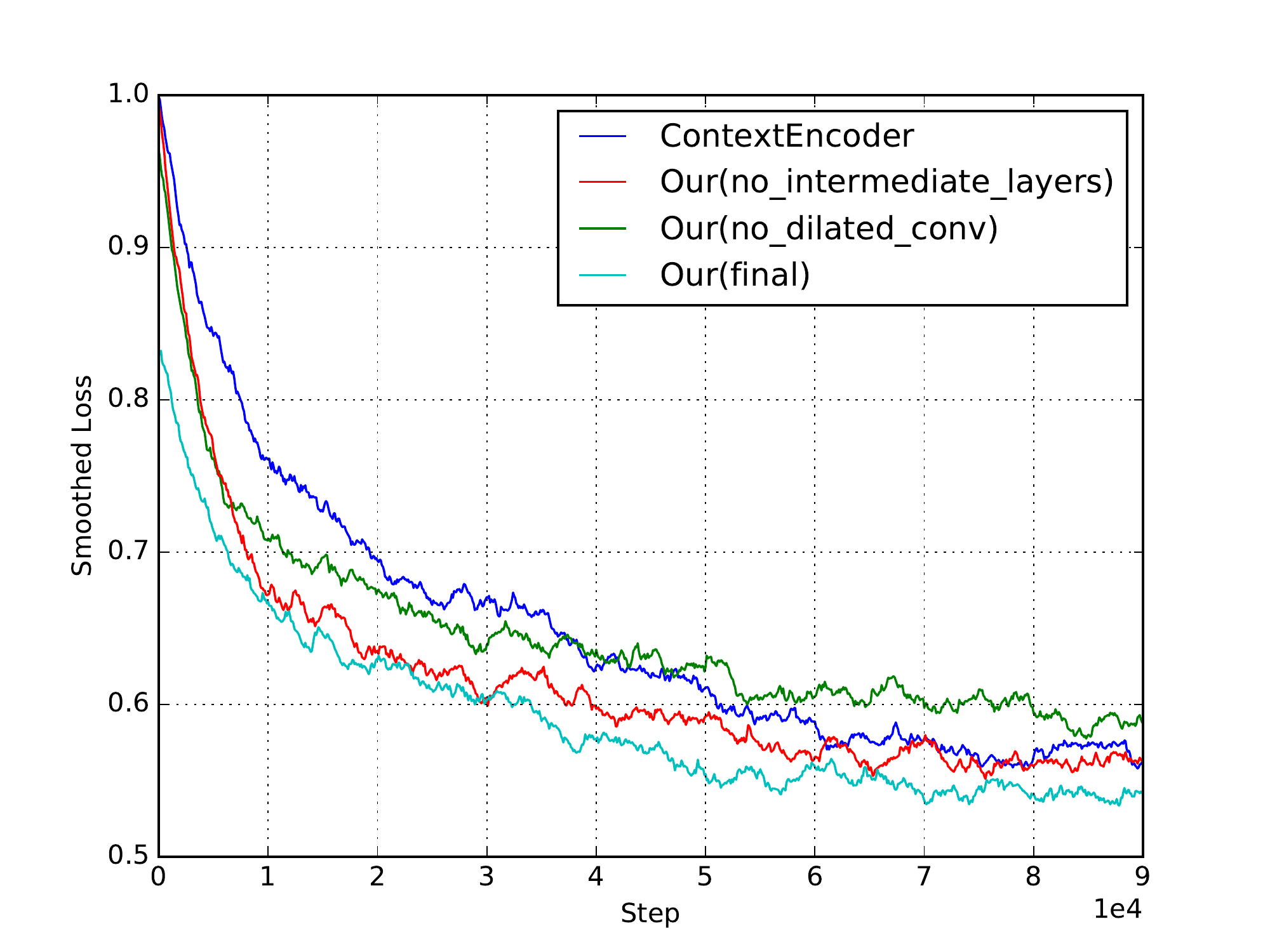}
\caption{Training loss curves of different generator structures. We train each model to regress the missing central region with reconstruction loss only on Paris StreetView.}
\label{Generator Compare}
\end{figure}
\textbf{Training Procedure.}
Instead of training the model with the joint loss all together directly, we add them gradually. First, we train the generator with reconstruction loss to synthesize coarse contents. Then adversarial loss and perceptual loss are added to further refine the results. We set learning rate with polynomial decay from $10^{-3}$ to $10^{-6}$.

\textbf{Hyperparameter.} During training, it is extremely important to fine-tune the hyperparameters so that we can make a good balance among different losses. In backpropagation, it is gradient values that make sense rather than loss values. Thus we propose a strategy to effectively select proper hyperparameters, which first trains the model for several steps to gain their gradients, then sets hyperparameters according to their relative magnitude. Choosing hyperparameters based on gradients enables us to know the exact effects of different losses.

\subsection{Comparison of Generator Structures}

As shown in Figure~\ref{Generator Compare}, dilated convolution layers can improve performance a lot, because it helps increase receptive field, letting generator see a 'larger' area.  Intermediate layers combine multi-scale information, and the connection between low layers and high layers can ease the problem of gradient vanish, thus we also gain performance improvement. Compared with the generator of Context Encoder, our generator achieves better results with less parameters, demonstrating the superiority of our generator structure.

\subsection{Effect of Original Image Context}
The aim of image completion is to synthesize contents that fit the original image context well. We conduct an experiment to show the effect of image context, which employs two different input settings for the input of discriminator: replace unmasked regions with original pixels or no replace. As shown in Figure~\ref{Replace Compare}, when utilizing the original image context, the synthesized contents are more continuous on the boundary and the image completion results are more natural. 

\subsection{Analysis and Comparison of Perceptual Features}
Before using perceptual features, we need to know the exact effects of different levels of perceptual features. We make detailed comparison and analysis. As shown in Figure~\ref{Perceptual comparison}, when using perceptual features extracted from $conv1\_2$ and $conv2\_2$, the inpainting results are still blurry. This is because $conv1\_2$ and $conv2\_2$ only capture low-level information without capturing some characteristics of structures and shapes. However, (d)(e)(f) have clear face and building structures, showing that higher layers such as $conv3\_2$, $conv4\_2$ and $conv5\_2$ can capture some semantic information which helps generate details. To combine the benefits of them, we use the ReLU output of layers $conv3\_2$, $conv4\_2$ and $conv5\_2$ in VGG16 as perceptual loss in our experiments, which achieves the best performance Figure~\ref{Perceptual comparison: f}. 
\subsection{Qualitative Comparisons}
We first compare our method with Photoshop Content Aware Fill (CAF, based on PatchMatch \cite{barnes2009patchmatch}) and Generative Face Completion (GFC, \cite{li2017generative}) on CelebA (Figure~\ref{CelebA Compare}). We do not compare with Context Encoder (CE, \cite{pathak2016context}) here, because CE can only work on the same masks used in training, while our evaluations here focus on cases with random mask size and location. Traditional methods such as CAF, cannot handle highly specific and complex structures such as faces, because every face has different structure and copy-and-paste strategy will not work. GFC can synthesize some key components such as nose, but it will fail if the faces are not aligned, such as the first row and the last row. Moreover, as shown in Figure~\ref{CelebA Compare: d} and \ref{CelebA Compare: e}, the synthesized components is not consistent with surrounding regions, which needs some post-processing methods to eliminate. By contrast, our method is not sensitive to position and face does not need to be aligned, because both the mask size and location is random during our training, and our network is a FCN, which has the property of translation invariance. 

We then compare our method with CAF and CE on Paris StreetView (Figure~\ref{Paris Compare}). Our method can work well on any mask location. Since CE can only work on specific masks, we compare the results on the central region. CAF can work well on texture structures, such as the third row in Figure~\ref{Paris Compare}, but it does not capture the semantic information. Compared with CE, our results are consistent on the boundary and have fine-grained details.

Figure~\ref{CelebA Compare: d}, \ref{CelebA Compare: f} and Figure~\ref{Paris Compare: d}, \ref{Paris Compare: e} are all trained with reconstruction loss and adversarial loss. The results show the superiority of our network structure, which can synthesize more consistent and realistic contents. Comparison of Figure~\ref{CelebA Compare: f}, \ref{CelebA Compare: g} and Figure~\ref{comparision: e}, \ref{comparision: f} describe the advantages of perceptual loss, which can help synthesis details.

\begin{figure}[t]
\centering
\subfigure[]{
\begin{minipage}[b]{0.11\textwidth}
\includegraphics[width=1\textwidth]{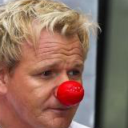} \\ 
\includegraphics[width=1\textwidth]{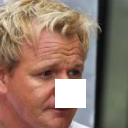} \\
\includegraphics[width=1\textwidth]{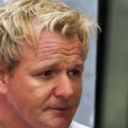}
\end{minipage}
\label{Object Removal: a}
}
\hspace{-0.015\textwidth}
\subfigure[]{
\begin{minipage}[b]{0.11\textwidth}
\includegraphics[width=1\textwidth]{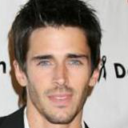} \\ 
\includegraphics[width=1\textwidth]{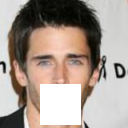} \\
\includegraphics[width=1\textwidth]{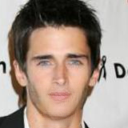}
\end{minipage}
\label{Object Removal: b}
}
\hspace{-0.015\textwidth}
\subfigure[]{
\begin{minipage}[b]{0.11\textwidth}
\includegraphics[width=1\textwidth]{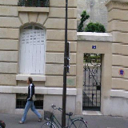} \\ 
\includegraphics[width=1\textwidth]{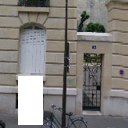} \\
\includegraphics[width=1\textwidth]{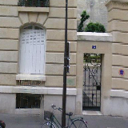} 
\end{minipage}
\label{Object Removal: c}
}
\hspace{-0.015\textwidth}
\subfigure[]{
\begin{minipage}[b]{0.11\textwidth}
\includegraphics[width=1\textwidth]{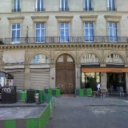} \\ 
\includegraphics[width=1\textwidth]{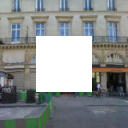} \\
\includegraphics[width=1\textwidth]{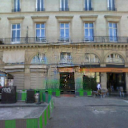} 
\end{minipage}
\label{Object Removal: d}
}
\caption{Object removal examples on CelebA and Paris Street.}
\label{Object Removal}
\end{figure}
\subsection{Quantitative Comparisons} We compare our method with baseline methods Context Encoder \cite{pathak2016context} on Paris StreetView dataset. As shown in Table \ref{Quantitative Comparisons}, our method can achieve better results. Our method has nearly no performance decrease in the random setting, while Context Encoder decreases a lot. The results show that our method is capable for image completion with random mask location.  

Compared with the results of $L_r+L_a$, our final results has comparable performance, because semantic inpainting is not for recovering the original images, but try to fill in the missing regions with realistic contents, as shown in Figure~\ref{CelebA Compare}, Figure~\ref{Paris Compare} and Figure~\ref{Object Removal}. Therefore the quantitative evaluation may not be an effective metric. 

\begin{table}[!htbp]\small
\centering
\caption{Comparisons with Context Encoder \cite{pathak2016context} on Paris StreetView Dataset. Mask size is $56\times56$, up/down are results of center/random region completion.}
\label{Quantitative Comparisons}
\begin{tabular}{lccc}
\hline
Methods                         & Mean $L_1$ Loss & Mean $L_2$ Loss &     PSNR       \\ \hline
\multirow{2}{*}{Context Encoder}&      0.1487     &     0.0545      &    20.4878dB   \\ \cline{2-4} 
                                &      0.2205     &     0.1005      &    17.3369dB   \\ \hline
\multirow{2}{*}{Ours($L_r+L_a$)}&      0.1320     &     0.0456      &    21.3919dB   \\ \cline{2-4} 
                                &      0.1338     &     0.0472      &    21.3863dB   \\ \hline
\multirow{2}{*}{Ours(final)}    &      0.1310     &     0.0436      &    21.5338dB   \\ \cline{2-4} 
                                &      0.1334     &     0.0460      &    21.5059dB   \\ \hline
\end{tabular}
\end{table}

\subsection{Object Removal}
Our algorithm can also be employed to remove unwanted objects, some examples are shown in Figure~\ref{Object Removal}. The red object in the nose of \ref{Object Removal: a}, the beard of \ref{Object Removal: b}, the people in \ref{Object Removal: c}, the brown door of \ref{Object Removal: d} are removed, but the results still look natural. In Figure \ref{Object Removal: d}, there are red banners on top of the two rooms in the right corner, therefore the synthesized center door also has a red banner, which shows the reasoning ability of our algorithm.

\subsection{Limitations}

\begin{figure}[t]
\centering
\subfigure[]{
\begin{minipage}[t]{0.11\textwidth}
\includegraphics[width=1\textwidth]{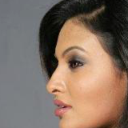} \\
\includegraphics[width=1\textwidth]{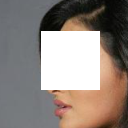} \\
\includegraphics[width=1\textwidth]{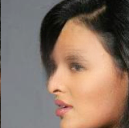}
\end{minipage}
\label{Failure Cases: a}
}
\hspace{-0.015\textwidth}
\subfigure[]{
\begin{minipage}[t]{0.11\textwidth}
\includegraphics[width=1\textwidth]{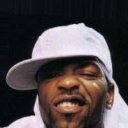} \\
\includegraphics[width=1\textwidth]{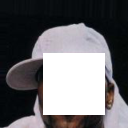} \\
\includegraphics[width=1\textwidth]{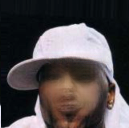}
\end{minipage}
\label{Failure Cases: b}
}
\hspace{-0.015\textwidth}
\subfigure[]{
\begin{minipage}[t]{0.11\textwidth}
\includegraphics[width=1\textwidth]{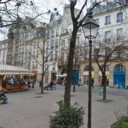} \\
\includegraphics[width=1\textwidth]{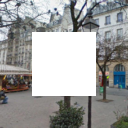} \\
\includegraphics[width=1\textwidth]{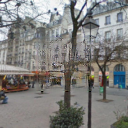} 
\end{minipage}
\label{Failure Cases: c}
}
\hspace{-0.015\textwidth}
\subfigure[]{
\begin{minipage}[t]{0.11\textwidth}
\includegraphics[width=1\textwidth]{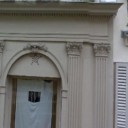} \\
\includegraphics[width=1\textwidth]{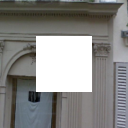} \\
\includegraphics[width=1\textwidth]{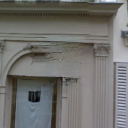}
\end{minipage}
\label{Failure Cases: d}
}
\caption{Failure cases of our approach.} 
\label{Failure Cases}
\end{figure}

Although our approach could synthesize semantically consistent results with fine-grained details and generate novel contents that do not appear in the dataset, it could fail if there are few or no scenes in the dataset.  Figure~\ref{Failure Cases} shows several failure cases. CelebA  has very few profile images, so our algorithm does not handle Figure~\ref{Failure Cases: a} well. Figure~\ref{Failure Cases: b} has a hat and the whole face is masked with nearly no pixels of skin color left. The algorithm does not find that the tree in Figure~\ref{Failure Cases: c} is crooked and does not utilize the symmetry property to complete arch structure of Figure~\ref{Failure Cases: d}, because most images in Paris StreetView are buildings and streets. 
\section{Conclusion}
In this paper, we present a novel approach for semantically consistent image completion. We make a detailed analysis and comparison on the effects of different losses and different levels of perceptual features and show that there exist complementarity between adversarial training and perceptual features. We design a novel lightweight generator structure and show the superiority of our method on two benchmark datasets.
{\small
\bibliographystyle{ieee}
\bibliography{egbib}
}

\clearpage
\section*{Supplementary Material}

\begin{figure}[h]
\centering
\subfigure[Original]{
\begin{minipage}[b]{0.14\textwidth}
\includegraphics[width=1\textwidth]{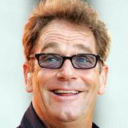}  \\ 
\includegraphics[width=1\textwidth]{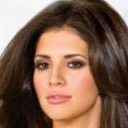}  \\
\includegraphics[width=1\textwidth]{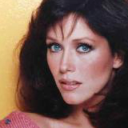}  \\ 
\includegraphics[width=1\textwidth]{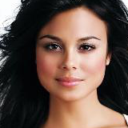}  \\
\includegraphics[width=1\textwidth]{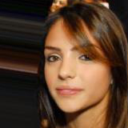}  \\ 
\includegraphics[width=1\textwidth]{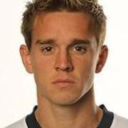}  \\
\includegraphics[width=1\textwidth]{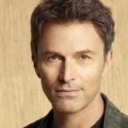}  \\ 
\includegraphics[width=1\textwidth]{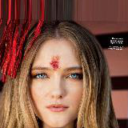}    
\end{minipage}
}
\hspace{-0.015\textwidth}
\subfigure[Masked]{
\begin{minipage}[b]{0.14\textwidth}
\includegraphics[width=1\textwidth]{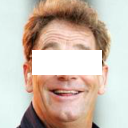} \\
\includegraphics[width=1\textwidth]{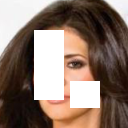} \\
\includegraphics[width=1\textwidth]{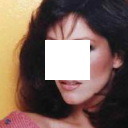} \\
\includegraphics[width=1\textwidth]{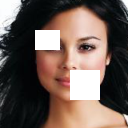} \\
\includegraphics[width=1\textwidth]{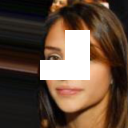} \\
\includegraphics[width=1\textwidth]{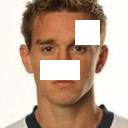} \\
\includegraphics[width=1\textwidth]{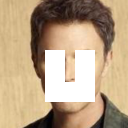} \\
\includegraphics[width=1\textwidth]{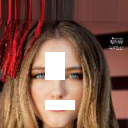}  
\end{minipage}
}
\hspace{-0.015\textwidth}
\subfigure[Results]{
\begin{minipage}[b]{0.14\textwidth}
\includegraphics[width=1\textwidth]{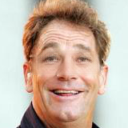} \\
\includegraphics[width=1\textwidth]{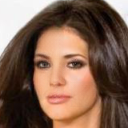} \\
\includegraphics[width=1\textwidth]{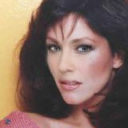} \\
\includegraphics[width=1\textwidth]{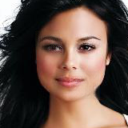} \\
\includegraphics[width=1\textwidth]{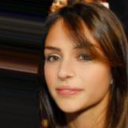} \\
\includegraphics[width=1\textwidth]{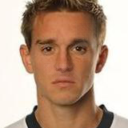} \\
\includegraphics[width=1\textwidth]{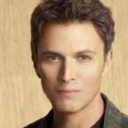} \\
\includegraphics[width=1\textwidth]{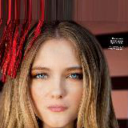} 

\end{minipage}
}
\caption{Results on CelebA Face Dataset.}
\end{figure}

\begin{figure}[b]
\centering
\subfigure[Original]{
\begin{minipage}[b]{0.14\textwidth}
\includegraphics[width=1\textwidth]{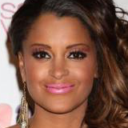}  \\ 
\includegraphics[width=1\textwidth]{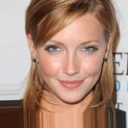}  \\
\includegraphics[width=1\textwidth]{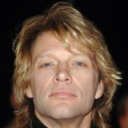}  \\ 
\includegraphics[width=1\textwidth]{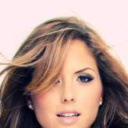}  \\
\includegraphics[width=1\textwidth]{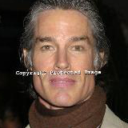}  \\ 
\includegraphics[width=1\textwidth]{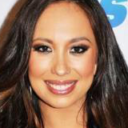}  \\
\includegraphics[width=1\textwidth]{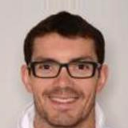}  \\ 
\includegraphics[width=1\textwidth]{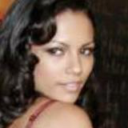}    
\end{minipage}
}
\hspace{-0.015\textwidth}
\subfigure[Masked]{
\begin{minipage}[b]{0.14\textwidth}
\includegraphics[width=1\textwidth]{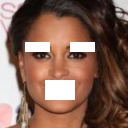} \\
\includegraphics[width=1\textwidth]{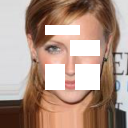} \\
\includegraphics[width=1\textwidth]{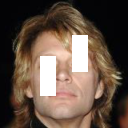} \\
\includegraphics[width=1\textwidth]{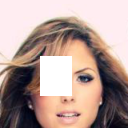} \\
\includegraphics[width=1\textwidth]{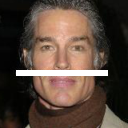} \\
\includegraphics[width=1\textwidth]{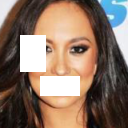} \\
\includegraphics[width=1\textwidth]{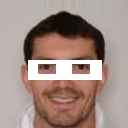} \\
\includegraphics[width=1\textwidth]{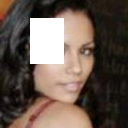}  
\end{minipage}
}
\hspace{-0.015\textwidth}
\subfigure[Results]{
\begin{minipage}[b]{0.14\textwidth}
\includegraphics[width=1\textwidth]{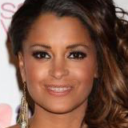} \\
\includegraphics[width=1\textwidth]{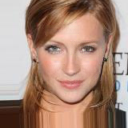} \\
\includegraphics[width=1\textwidth]{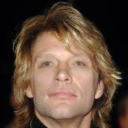} \\
\includegraphics[width=1\textwidth]{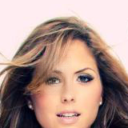} \\
\includegraphics[width=1\textwidth]{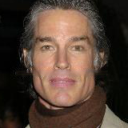} \\
\includegraphics[width=1\textwidth]{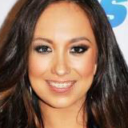} \\
\includegraphics[width=1\textwidth]{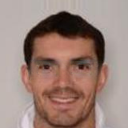} \\
\includegraphics[width=1\textwidth]{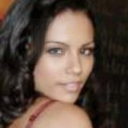} 

\end{minipage}
}
\caption{Results on CelebA Face Dataset.}
\end{figure}

\begin{figure}[t]
\centering
\subfigure[Original]{
\begin{minipage}[b]{0.144\textwidth}
\includegraphics[width=1\textwidth]{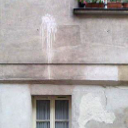}  \\ 
\includegraphics[width=1\textwidth]{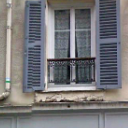}  \\
\includegraphics[width=1\textwidth]{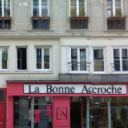}  \\ 
\includegraphics[width=1\textwidth]{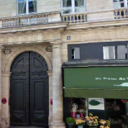}  \\
\includegraphics[width=1\textwidth]{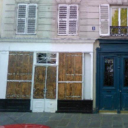}  \\ 
\includegraphics[width=1\textwidth]{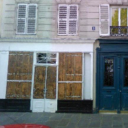}  \\
\includegraphics[width=1\textwidth]{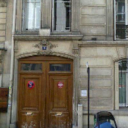}  \\ 
\includegraphics[width=1\textwidth]{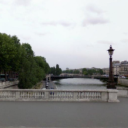} 
\end{minipage}
}
\hspace{-0.015\textwidth}
\subfigure[Masked]{
\begin{minipage}[b]{0.144\textwidth}
\includegraphics[width=1\textwidth]{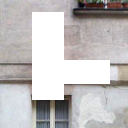} \\
\includegraphics[width=1\textwidth]{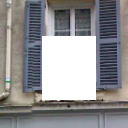} \\
\includegraphics[width=1\textwidth]{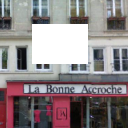} \\
\includegraphics[width=1\textwidth]{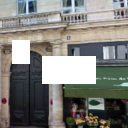} \\
\includegraphics[width=1\textwidth]{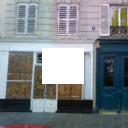} \\
\includegraphics[width=1\textwidth]{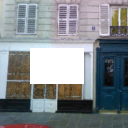} \\
\includegraphics[width=1\textwidth]{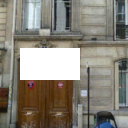} \\
\includegraphics[width=1\textwidth]{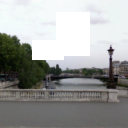} 
\end{minipage}
}
\hspace{-0.015\textwidth}
\subfigure[Results]{
\begin{minipage}[b]{0.144\textwidth}
\includegraphics[width=1\textwidth]{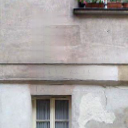} \\
\includegraphics[width=1\textwidth]{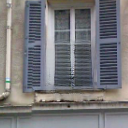} \\
\includegraphics[width=1\textwidth]{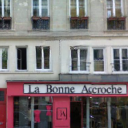} \\
\includegraphics[width=1\textwidth]{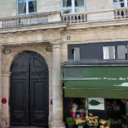} \\
\includegraphics[width=1\textwidth]{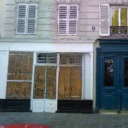} \\
\includegraphics[width=1\textwidth]{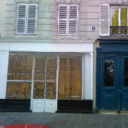} \\
\includegraphics[width=1\textwidth]{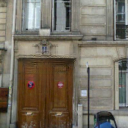} \\
\includegraphics[width=1\textwidth]{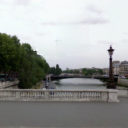} 

\end{minipage}
}
\caption{Results On Paris StreetView Dataset.}
\end{figure}

\end{document}